# Sensing-based Robustness Challenges in Agricultural Robotic Harvesting

C. Beldek
School of Mechanical, Materials, Mechatronic, and Biomedical Engineering
University of Wollongong
Wollongong, NSW, Australia
cbeldek@uow.edu.au

J. Cunningham
School of Mechanical, Materials, Mechatronic, and Biomedical Engineering
University of Wollongong
Wollongong, NSW, Australia

M. Aydin
School of Mechanical, Materials, Mechatronic, and Biomedical Engineering
University of Wollongong
Wollongong, NSW, Australia

E. Sariyildiz
School of Mechanical, Materials, Mechatronic, and Biomedical Engineering
University of Wollongong
Wollongong, NSW, Australia

S. L. Phung
School of Electrical, Computer and Telecommunications Engineering
University of Wollongong
Wollongong, NSW, Australia

G. Alici
School of Mechanical, Materials, Mechatronic, and Biomedical Engineering
University of Wollongong
Wollongong, NSW, Australia

*Abstract* —This paper presents the challenges agricultural robotic harvesters face in detecting and localising fruits under various environmental disturbances. In controlled laboratory settings, both the traditional HSV (Hue Saturation Value) transformation and the YOLOv8 (You Only Look Once) deep learning model were employed. However, only YOLOv8 was utilised in outdoor experiments, as the HSV transformation was not capable of accurately drawing fruit contours. Experiments include ten distinct fruit patterns with six apples and six oranges. A grid structure for homography (perspective) transformation was employed to convert detected midpoints into 3D world coordinates. The experiments evaluated detection and localisation under varying lighting and background disturbances, revealing accurate performance indoors, but significant challenges outdoors. Our results show that indoor experiments using YOLOv8 achieved 100% detection accuracy, while outdoor conditions decreased performance, with an average accuracy of 69.15% for YOLOv8 under direct sunlight. The study demonstrates that real-world applications reveal significant limitations due to changing lighting, background disturbances, and colour and shape variability. These findings underscore the need for further refinement of algorithms and sensors to enhance the robustness of robotic harvesters for agricultural use.

## I. Introduction

As the world population is estimated to reach 9.7 billion by 2050, providing sufficient food for everyone is becoming more challenging each day [1]. According to the sustainability goals of the UN (United Nations), labour-intensive agricultural practices have been highlighted as one of the main reasons for the lack of food access [2, 3]. To present a viable solution to meet rising food demand, agricultural production steps from sowing to harvesting must become more sustainable and resilient [4].

Despite many improvements in modern agriculture, harvesting remains one of the most manual agricultural practices, relying on human skill and being prone to labour-related issues such as high costs and crop loss. It is also slow and inefficient, thus failing to provide a sustainable solution for the future of harvesting [5]. Bulk harvesting strategies, which are capable of efficiently harvesting cereals, olives, and potatoes, are available on the market [6, 7]. However, when it comes to specialty crops, such as most fruits, vegetables, and nursery products [8], bulk harvesting suffers from a lack of environmental adaptability and delicate harvesting strategies [9]. Due to their destructive and quality- and maturity-independent picking approach, the crops that can be harvested using bulk methods are limited [10]. Moreover, most bulk harvesters on the market still require a human operator to execute tasks in unpredictable environmental settings. Thus, neither current manual nor bulk harvesting practices can completely address labour-intensity issues in the agriculture industry.

To pick specialty crops effectively and quickly without human contribution, robotic systems have recently become a viable option [11]. Robotic harvesters are custom agricultural systems typically equipped with robotic arms, mobile platforms, sensors, and end-effectors [12-14]. Since studies dating back to the 1980s [15, 16], manipulator designs [17, 18], grasping strategies [19–21], motion control techniques [22, 23], and sensing methods [24, 25] for robotic pickers have continued to evolve to adapt to current technological trends. However, unlike traditional industrial robotic manipulators that work in well-structured indoor environments, harvesting robots are exposed to various environmental effects. Robustness and adaptability are key requirements for robotic systems interacting with environments [26–29]. Hence, even with promising advancements in robotic harvesters, achieving a commercially viable robotic picker remains a challenging problem due to environmental disturbances [30, 31].

Sensing is vital for detecting and locating target crops to guide real harvesting practices before picking. With improvements in cameras and detection methods, and their increasing affordability and applicability, vision-based sensing strategies have assumed a significant role in robotic harvesting [32-34]. For example, ML (Machine Learning) and DL (Deep Learning)-based vision algorithms have been widely used with robotic pickers for the detection and localisation process due to their ability to accurately detect agricultural targets [35, 36]. ML- and DL-based algorithms are generally capable of meeting the high accuracy and adaptability requirements for robotic harvesting. However, these techniques exhibit sensitivity that negatively impacts the robustness of detection success in outdoor environments [37]. Therefore, to present an accurate, applicable, and robust detection and localisation solution for robotic harvesting, current methods must be examined alongside state-of-the-art sensing challenges.

This work assesses the detection performance of HSV transformation and the YOLOv8 model for agricultural robotic harvesting under environmental challenges, including changing lighting conditions, background disturbances, and colour and shape variability. To demonstrate how these challenges affect the detection models of the robotic harvester, various experiments were performed to detect the location of apples and oranges in both indoor and outdoor environments using different detectors. Several studies addressing HSV transformation have emphasised its poor recognition success under varying illumination and background disturbances. The experimental results aligned with the approximately 0% recognition success of HSV outdoors and showed an inverse relationship between detection performance and background disturbances, as seen in [38] and [39]. A deep learning model YOLOv8 exhibited significantly higher accuracy than a colour-based model HSV in outdoor environments. This study confirms that the interaction between detection methods, environmental challenges, and crop types in robotic harvesting can vary significantly due to the unpredictable nature of agricultural scenes, often influenced by disturbances.

The remainder of the paper is organised as follows. Section II explains the detection and localisation methods for recognising crops. Section III describes the conducted experiments. Section IV provides an evaluation of detection and localisation across various environmental conditions. Finally, Section V provides the study's conclusion.

## II. Detection and Localisation Methods

### A. Target Detection

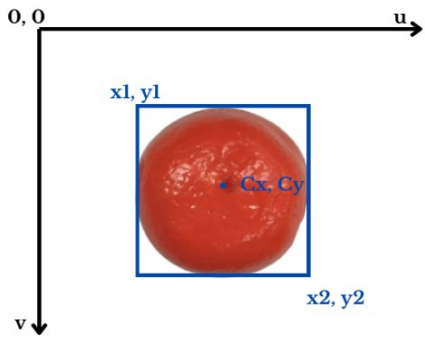

Figure 1 Midpoint calculation

The fruit detection process is a precursor to localisation, which calculates the picking point position of the target. Typically, the picking point can be selected as the midpoint of spherical fruits, as shown in Fig. 1. If the fruit is not spherical, picking points might be located on the fruit's peduncles. In the image frame, the picking points of the target fruit ($C_x$ and $C_y$) are defined as follows:

$$C_x = (x_1 + x_2)/2 \text{ and } C_y = (y_1 + y_2)/2 \quad (1)$$

where $x_1$ and $y_1$ represent the coordinates of the bounding box's top-left corner; and $x_2$ and $y_2$ represent the coordinates of the bottom-right corner.

The HSV transformation, which considers only the colour feature, is employed to detect the target fruit using the OpenCV library. Sequentially, the colour image is masked, contours are obtained and picking points are defined in HSV colour space, as shown in Eq. (1).

Unlike traditional methods, deep learning models mostly utilise multi-features to recognise target fruit in the image frame. As a single-stage deep learning model, pretrained YOLOv8 using the COCO dataset was employed in indoor and outdoor experiments [40, 41]. A similar midpoint calculation method is applied to the bounding boxes, which are the output of the pre-trained YOLOv8 model.

### B. Localisation

The image projection model that allows transforming pixel points to world coordinates can be obtained as shown in Eq. (2).

$$\begin{bmatrix} u \\ v \\ w \end{bmatrix} = \begin{bmatrix} \mathbf{K} & 0 \\ 0 & 1 \end{bmatrix} \begin{bmatrix} \mathbf{R} & \mathbf{t} \\ 0 & 1 \end{bmatrix} \begin{bmatrix} X \\ Y \\ Z \\ 1 \end{bmatrix} \quad (2)$$

where $u, v$ and $w$ are the horizontal, vertical and depth pixel coordinates in the image frame, respectively; $X, Y$ and $Z$ are the horizontal, vertical and depth pixel coordinates in the world coordinate frame, respectively. $\mathbf{K} = \begin{bmatrix} f_x & 0 & c_x \\ 0 & f_y & c_y \\ 0 & 0 & 1 \end{bmatrix}$ is the intrinsic camera matrix in which $f_x$ and $f_y$ are focal lengths for each axis in pixels, and $c_x$ and $c_x$ are the centre points in pixels. $\mathbf{R} \mid \mathbf{t} = \begin{bmatrix} r_{11} & r_{12} & r_{13} & t_1 \\ r_{21} & r_{22} & r_{23} & t_2 \\ r_{31} & r_{32} & r_{33} & t_3 \end{bmatrix}$ is the transformation matrix that combines the rotation matrix $\mathbf{R}$ and translation vector $\mathbf{t}$ in which; $r_{ij}$ represents the $i$-th and $j$-th term of $\mathbf{R}$, and $t_i$ represents the $i$-th term of the $\mathbf{t}$. To estimate the pose of the target fruit, homography transformation has been employed. To this end, the following camera model is used:

$$\mathbf{C} = \mathbf{K} \, \mathbf{R} \mid t \quad (3)$$

where $\mathbf{C}$ is the camera transformation matrix which encompasses both intrinsic and extrinsic parameters to transform the pixel coordinates to the world coordinates.

The homography matrix that can transform a point from the world coordinate frame to the image coordinate frame can be obtained from $\mathbf{C}$. This assumes that the world coordinate frame is flat and lies in the plane $(X, Y, 0)$. Finally, the unknown homography matrix can be addressed as a Homogeneous Linear Least Squares problem and solved using Direct Linear Transformation (DLT) algorithm and Singular Value Decomposition (SVD), as detailed in [42].

## III. Experiments

In this study, a tripod was used to hold an Intel RealSense RGB415 camera, positioned downward at a height of approximately 61.5 cm. A grid made of two metal plates, each with holes equally spaced horizontally and vertically, was featured. This arrangement allowed for easy location determination relative to the origin (the top left corner of the grid) and facilitated the creation of random patterns. The experimental setup is shown in Fig. 2.

Since diverse lighting conditions [43, 44] and background disturbances [45, 46] impacted the accurate detection and localisation of agricultural products with a unique colour and shape, experiments were designed to address these challenges. To this end, experiments were conducted with ten random patterns of apples and oranges, with each pattern tested twice to account for the influence of background disturbances and colour and shape variability. Data from indoor experiments were collected in a laboratory workspace with constant lighting. Subsequent experiments were performed outdoors in two distinct locations to demonstrate environmental differences. The first location was a shaded area covered by trees, limiting sunlight exposure on the camera and fruits. The second site was an open area that was directly under the sun, subjecting the camera and fruits to full sunlight exposure. Both experiments took place in the afternoon at the same time but on separate days.

*A. Indoor Experiments*

The indoor experiments assessed object detection and midpoint localisation for apples and oranges, both with and without disturbances. The methods used in the indoor experiments were HSV transformation and YOLOv8, as both were applicable under controlled lighting, as shown in Figs. 3a to 3h. HSV transformation-based fruit contours were drawn to identify the midpoint of the target fruit (see Figs. 3b, 3d, 3f, and 3h). Conversely, the bounding box structure was employed to distinguish the target in YOLOv8. Figures 3a, 3b, 3e and 3f depict the process without disturbances, while Figures 3c, 3d, 3g and 3h show it with disturbances.

The detection accuracy of both methods was 100% without background disturbances in controlled settings. However, background disturbances led to misidentifications in YOLOv8 and inconsistent contour generation in HSV transformation. For instance, Fig. 3g illustrates how leaves were misidentified as objects by the object detection model. Additionally, in Fig. 3h, the similar colour of the fruit and leaves caused difficulty distinguishing between them, leading to contours being generated on the leaves on the fruit.

Contours were sometimes generated only on portions of the fruit during the HSV transformation, and subsequent frames frequently exhibited skips or delays in contour formation. These issues highlight the challenge of differentiating between objects with similar colours.

*B. Outdoor Experiments*

TABLE 1. OVERALL RESULTS OF INDOOR AND OUTDOOR EXPERIMENTS

| Crops | Metrics | With Disturbances | | | No Disturbances | | |
|---|---|---|---|---|---|---|---|
| | | Indoor | Shaded | Direct Sunlight | Indoor | Shaded | Direct Sunlight |
| Orange | X mean (cm) | 0.023 | 0.495 | 0.443 | 0.596 | 0.633 | 0.407 |
| Orange | Y mean (cm) | 0.339 | 1.681 | 1.182 | 0.339 | 1.79 | 1.240 |
| Orange | Detection (%) | 100 | 75 | 88.3 | 100 | 86.7 | 35 |
| Apple | X mean (cm) | 0.202 | 0.698 | 0.554 | 0.989 | 0.987 | 0.658 |
| Apple | Y mean (cm) | 0.384 | 1.414 | 1.269 | 0.778 | 1.768 | 1.32 |
| Apple | Detection (%) | 100 | 98.3 | 75 | 100 | 96.7 | 78.3 |

Apart from background disturbances and colour and shape variability, illumination variability was included in the experiment as another disturbance. HSV transformation method was oversensitive to illumination change. Therefore, only YOLOv8 was employed in outdoor experiments to address illumination variability issues, as can be seen in Figs. 4a to 4h. Midpoint localisation was achieved by calculating the midpoint of the bounding box around each fruit.

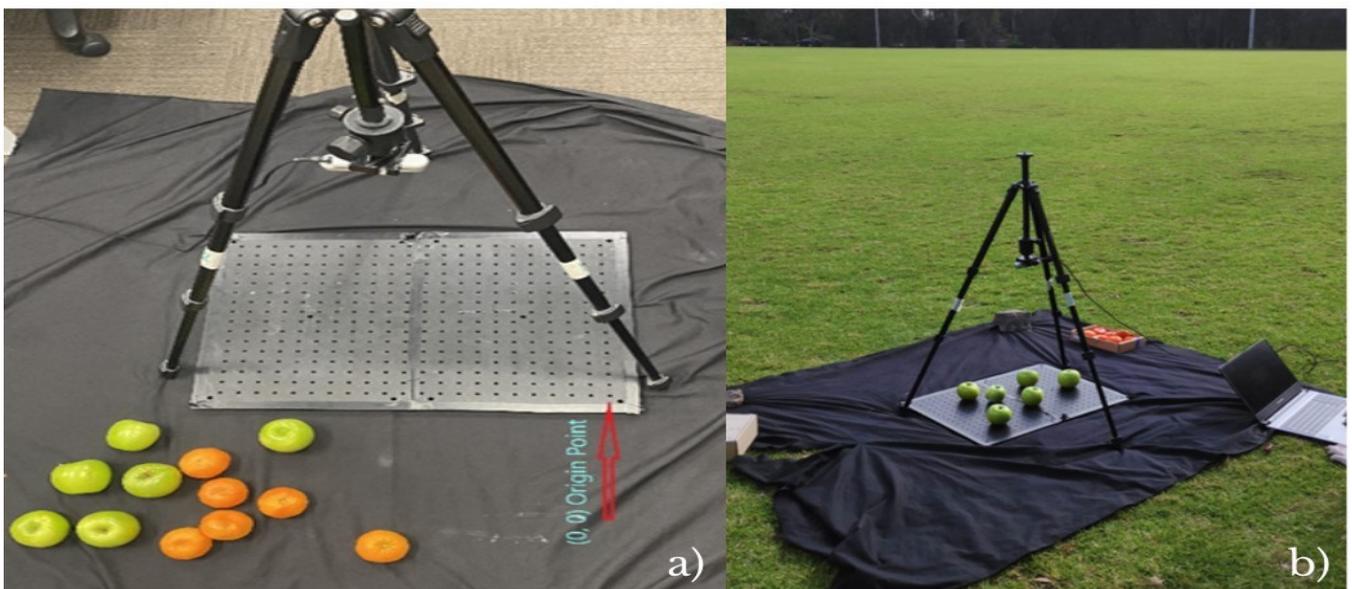

Figure 2: Experimental setup: (a) Indoor, with the origin point at the right corner of the grid structure, (b) outdoor.

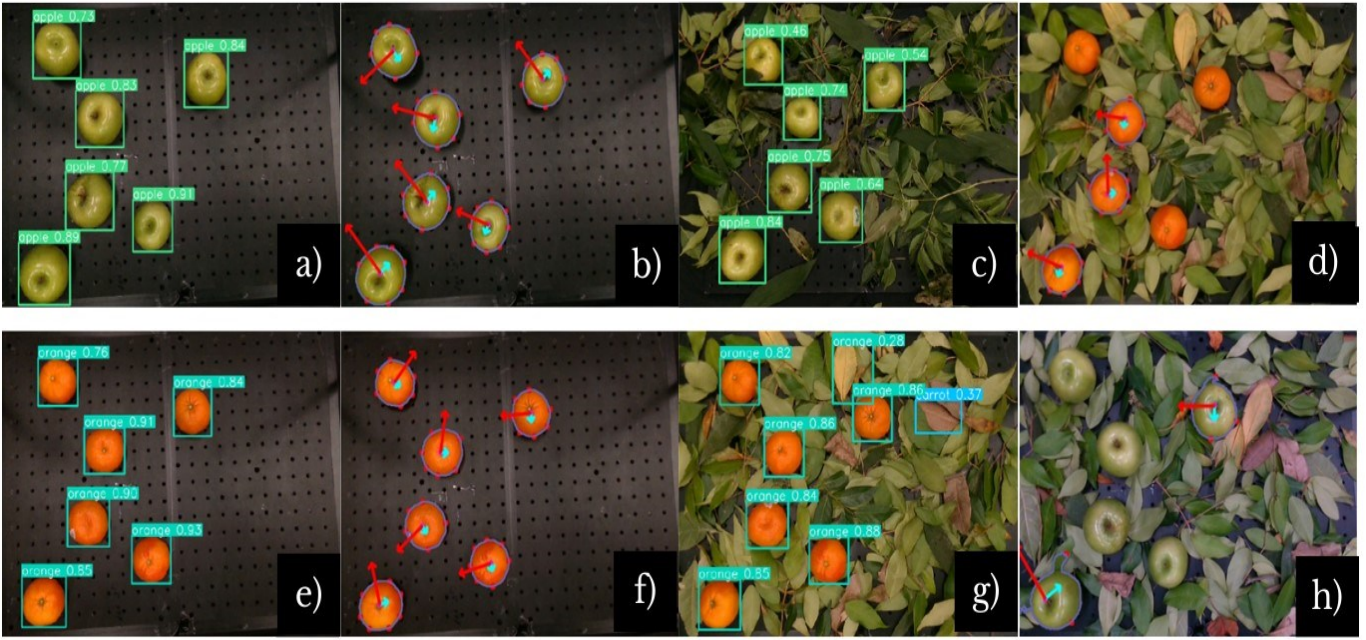

Figure 3: Indoor detection outputs: (a) indoor apple detection without background disturbances using YOLOv8, (b) indoor apple detection without background disturbances using HSV transformation, (c) indoor apple detection with background disturbances using YOLOv8, (d) indoor apple detection with background disturbances using HSV transformation, (e) indoor orange detection without background disturbances using YOLOv8, (f) indoor apple detection without background disturbances using HSV transformation, (g) indoor apple detection with background disturbances using YOLOv8, (h) indoor apple detection with background disturbances using HSV transformation.

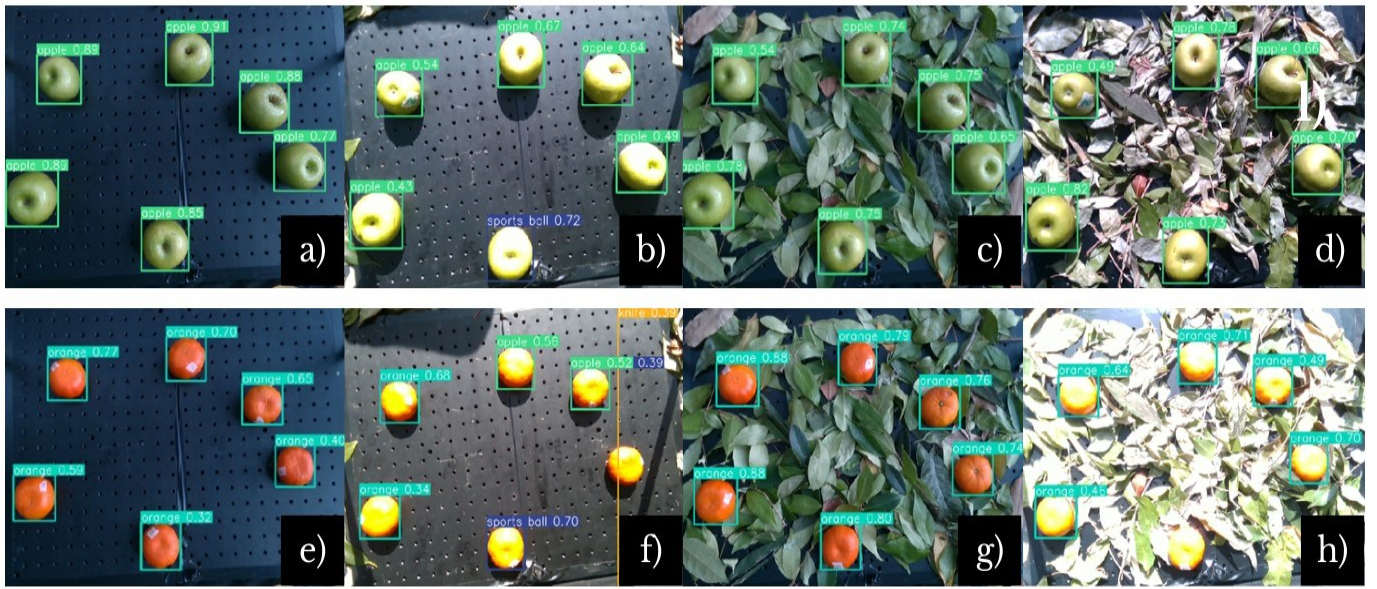

Figure 4: Outdoor detection outputs: (a) outdoor apple detection without background disturbances using YOLOv8, (b) outdoor apple detection without background disturbances using YOLOv8, (c) outdoor apple detection with background disturbances using YOLOv8, (d) outdoor apple detection with background disturbances using YOLOv8, (e) outdoor apple detection without background disturbances using YOLOv8, (f) outdoor apple detection without background disturbances using YOLOv8, (g) outdoor apple detection with background disturbances using YOLOv8, (h) outdoor apple detection with background disturbances using YOLOv8.

Under shaded lighting conditions, both with and without background disturbances, the detection rate was high, as demonstrated in Figs. 4a, 4c, 4e and 4g. Various repetitive misidentifications occurred in the shaded area: e.g., some oranges were detected as apples, and some apples were mistaken for sports balls due to shape similarity.

As shown in Figs. 4b, 4d, 4f and 4h, the open field experiments were conducted under direct sunlight at the same time on different days as in the shaded area. The fruit patterns observed under direct sunlight were the same as those in the shaded area, for easy comparison. The detection rate was lower than the shaded area's detection rates. Expose to direct sunlight results in shadow casting as illustrated in Fig. 4f, where the shadow was mistakenly detected as an object.

## IV. EVALUATION OF DETECTION AND LOCALISATION

The experimental results, illustrated in Fig. 3 and Fig. 4, involve the detection and localisation of six apples and six oranges. The varying sizes and colours of the fruits influence contour and bounding box generation, thereby changing midpoint localisation. Table 1 presents the true-positive detection rates and error values for the coordinates across ten patterns in indoor and outdoor environments, considering scenarios with and without background disturbances. Due to the inadequate performance of the HSV transformation under outdoor conditions, coordinates were derived from YOLOv8's bounding boxes only.

Under consistent indoor lighting, YOLOv8 achieved 100% detection accuracy, confirming its reliability in controlled environments. Similarly, the HSV transformation exhibited high accuracy due to the stability of pixel values under constant lighting. The average deviations for localisation in the X and Y axes were 0.453 cm and 0.460 cm, respectively, demonstrating precise localisation in the controlled laboratory settings.

The results of experiments varied significantly in outdoor environments. In shaded areas, the detection rate averaged 89.2%, while in open areas with direct sunlight, it dropped to 69.15%. Despite this, most detection rates still exceeded a 75% accuracy rate, with exceptions noted for oranges without disturbances under direct sunlight. Localisation accuracy in shaded areas had average deviations of 0.703 cm (X) and 1.663 cm (Y), whereas in open fields, the deviations were 0.516 cm (X) and 1.253 cm (Y). The difference between indoor and outdoor environments is primarily due to lighting variations. While YOLOv8 maintained high accuracy indoors, its outdoor performance was influenced by particularly sunlight exposure. Changing light conditions also lengthened the camera's automatic exposure time.

The effect of background disturbances increased the unpredictability of agricultural sensing due to the low colour contrast between the fruit and the background. This is reflected in the average detection rates of apples and oranges. Background disturbances negatively affected the detection performance of oranges. Under similar conditions, however, they had an unexpectedly positive effect on apple detection in shaded areas resulting in an improvement of less than 2%.

The limitations of the HSV transformation and YOLO object detectors in outdoor settings further emphasise the need for accurate and robust detection methods that can adapt to varying conditions. Although the results achieved are within the range of acceptable localisation errors for agricultural harvesting robots (typically 0–2 cm [47-49]), the robustness challenges faced by agricultural robotic sensing were highlighted in this study, particularly under natural lighting conditions.

To improve localisation performance in agricultural fields, advancements in sensing systems, multi-sensor fusion and innovations in vision sensors and lenses are essential. Incorporating image segmentation into detection models could also address localisation errors by providing more accurate contour generation, especially for asymmetrical or non-spherical crops. Especially, the problem of covered fruits by other tree parts can be solved by image segmentation techniques effectively instead of detection. Point cloud can help to 3D reconstruction process while segmenting target fruit pixels in the image frame to this end. Furthermore, non-sensing-based improvements, such as gripper design [36] and environmental structuring [50], can mitigate some challenges, but they do not fully resolve the underlying issues of sensing and localisation in unpredictable outdoor environments. To achieve robustness in real agricultural conditions, it is essential to refine algorithms and sensors further.

## V. Conclusion

This study highlights how environmental disturbances affect agricultural robotic harvesters. Agricultural environments are inherently unpredictable and prone to disturbances; therefore, accurate and robust detection and localisation are essential for successful robotic harvesting. A series of experiments were conducted in both indoor and outdoor environments to address various common sensing challenges, such as changing lighting conditions, background disturbances, and colour and shape variability. We employed HSV transformation and a pre-trained YOLOv8 detection model and used homography transformation for fruit pose estimation. Our results indicate that detection and localisation are most successful in controlled laboratory conditions, achieving around 100% accuracy. Colour-based detection methods are not effective in agricultural scenes for robotic harvesting, and detection and localisation errors tend to increase in shaded areas and under direct sunlight. Additionally, background disturbances and colour and shape variability contribute to increased detection and localisation errors.

The experimental results confirm the unpredictability and complexity of agricultural environments and their significant impact on robotic harvesting success. These findings are in line with the results reported in [51, 52]. The results are presented and assessed to provide a comprehensive view of the practical challenges faced by robotic pickers. Future research will focus on optimising detection and localisation methods and developing innovative sensing strategies for agricultural harvesting applications.